\documentclass{article}

\usepackage[preprint]{neurips_2022}

\usepackage{lineno,hyperref}
\usepackage{colortbl}
\usepackage{amsmath,amssymb,amsfonts}
\usepackage{algorithmic}
\usepackage{graphicx}
\usepackage{float}
\usepackage{eucal}

\usepackage{wrapfig}
\usepackage[utf8]{inputenc} 
\usepackage[T1]{fontenc}    
\usepackage{hyperref}       
\usepackage{url}            
\usepackage{booktabs}       
\usepackage{amsfonts}       
\usepackage{nicefrac}       
\usepackage{microtype}      
\usepackage{xcolor}         
\usepackage{latexsym}
\usepackage{amssymb}
\usepackage{amsmath}
\usepackage{booktabs}
\usepackage{enumerate}
\usepackage{graphicx}
\usepackage{subfigure}
\usepackage{xspace}
\usepackage{float}
\usepackage{bbm}
\usepackage{bm}
\usepackage{multirow}
\usepackage{booktabs}
\usepackage{color}
\usepackage{framed}
\usepackage{stfloats}
\usepackage{iitem}
\usepackage{amssymb}
\usepackage{pifont}
\usepackage{enumitem}

\AtBeginDocument{%
  \providecommand\BibTeX{{%
    \normalfont B\kern-0.5em{\scshape i\kern-0.25em b}\kern-0.8em\TeX}}}
\setcitestyle{square,numbers}
\hypersetup{colorlinks,linkcolor={blue},citecolor={blue},urlcolor={purple}}  

\title{Parameter-Efficient and Student-Friendly\\ Knowledge Distillation}

\author{%
    Jun Rao\textsuperscript{\rm 1}\thanks{Equal contributions from both authors. Work was done when Jun was interning at JD Explore Academy.}, Xv Meng\textsuperscript{\rm 1*}, Liang Ding\textsuperscript{\rm 2},  Shuhan Qi\textsuperscript{\rm 1,\rm 3}\thanks{Corresponding Author.}, Dacheng Tao\textsuperscript{\rm 2}  \\
    \textsuperscript{\rm 1}Harbin Institute of Technology, Shenzhen\\
    \textsuperscript{\rm 2}JD Explore Academy\quad
    \textsuperscript{\rm 3}Peng Cheng Laboratory \\
    \texttt{rao7jun@gmail.com,mxx0822@foxmail.com}\\
    \texttt{dingliang1@jd.com,shuhanqi@cs.hitsz.edu.cn,dacheng.tao@gmail.com}
}

\begin{document}

\maketitle
\begin{abstract}
  Knowledge distillation (KD) has been extensively employed to transfer the knowledge from a large teacher model to the smaller students, where the parameters of the teacher are fixed (or partially) during training.
  Recent studies show that this mode may cause difficulties in knowledge transfer due to the mismatched model capacities. 
  To alleviate the mismatch problem, teacher-student joint training methods, e.g., online distillation, have been proposed, but it always requires expensive computational cost. 
  In this paper, we present a parameter-efficient and student-friendly knowledge distillation method, namely PESF-KD, to achieve efficient and sufficient knowledge transfer by updating relatively few partial parameters. 
  Technically, we first mathematically formulate the mismatch as the sharpness gap between their predictive distributions, where we show such a gap can be narrowed with the appropriate smoothness of the soft label.
  Then, we introduce an adapter module for the teacher, and only update the adapter to obtain soft labels with appropriate smoothness.
  Experiments on a variety of benchmarks show that PESF-KD can significantly reduce the training cost while obtaining competitive results compared to advanced online distillation methods. Code will be released upon acceptance.
\end{abstract}

\section{Introduction}
With the continuous growth of data scale and computing resources, the scale of deep neural networks has rapidly increased, from millions of parameters to billions of parameters~\cite{9343374}. Although these huge models, such as BERT \cite{bert}, GPT-3~\cite{gpt3}, and CLIP \cite{DBLP:conf/icml/RadfordKHRGASAM21/clip}, have powerful multitasking capabilities, they cannot be deployed on some edge devices with limited computing resources, which further limits the environmentally friendly and efficient deployment of deep learning applications.

\begin{figure*}[t!]
    \subfigure[Vanilla KD]{
    \begin{minipage}[t]{0.31\textwidth}
           \centering
           \includegraphics[width=\textwidth]{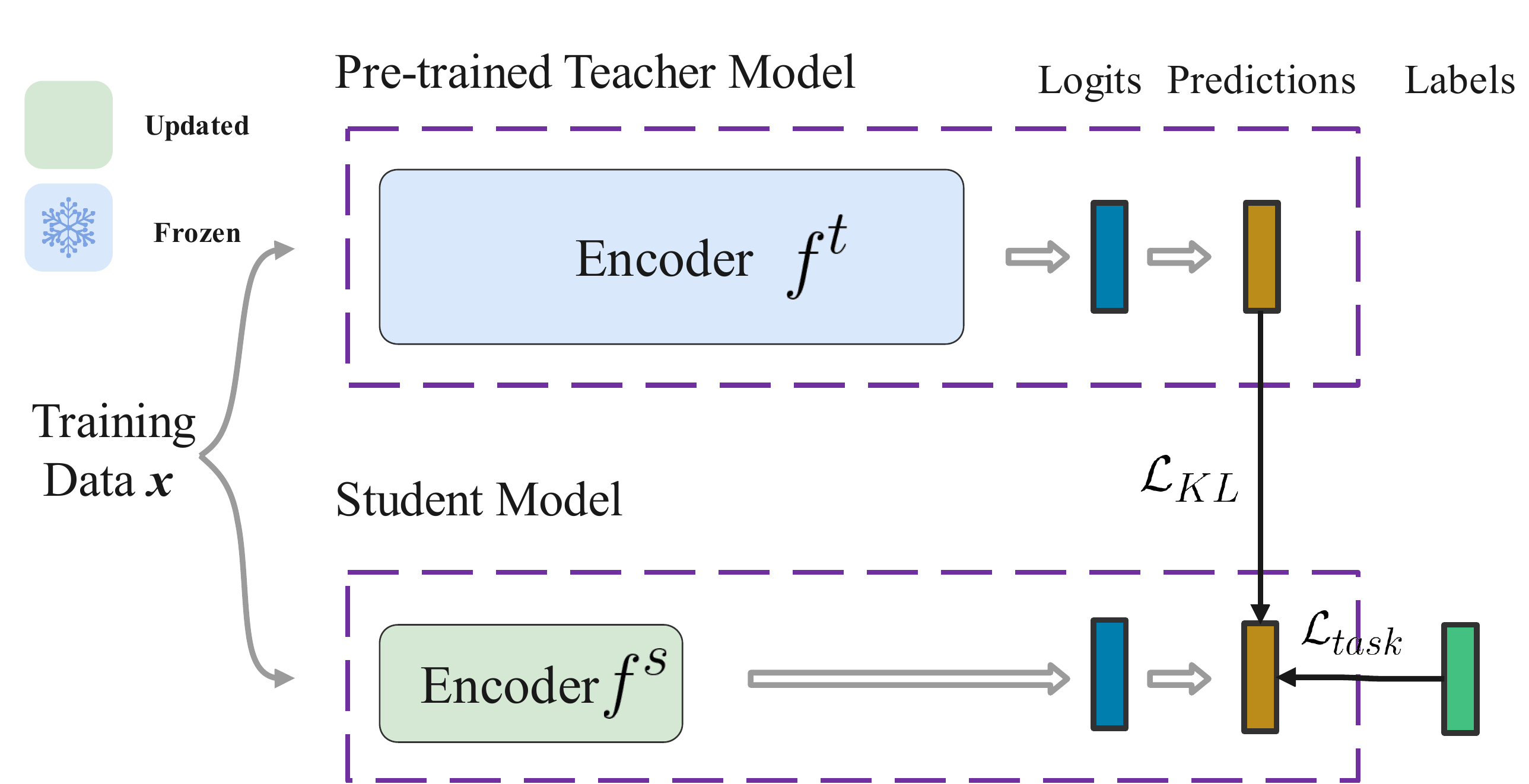}
            \label{fig:base_kd}
    \end{minipage}
    }
    \subfigure[KD with fine-tuned teacher]{
    \begin{minipage}[t]{0.31\textwidth}
            \centering
            \includegraphics[width=\textwidth]{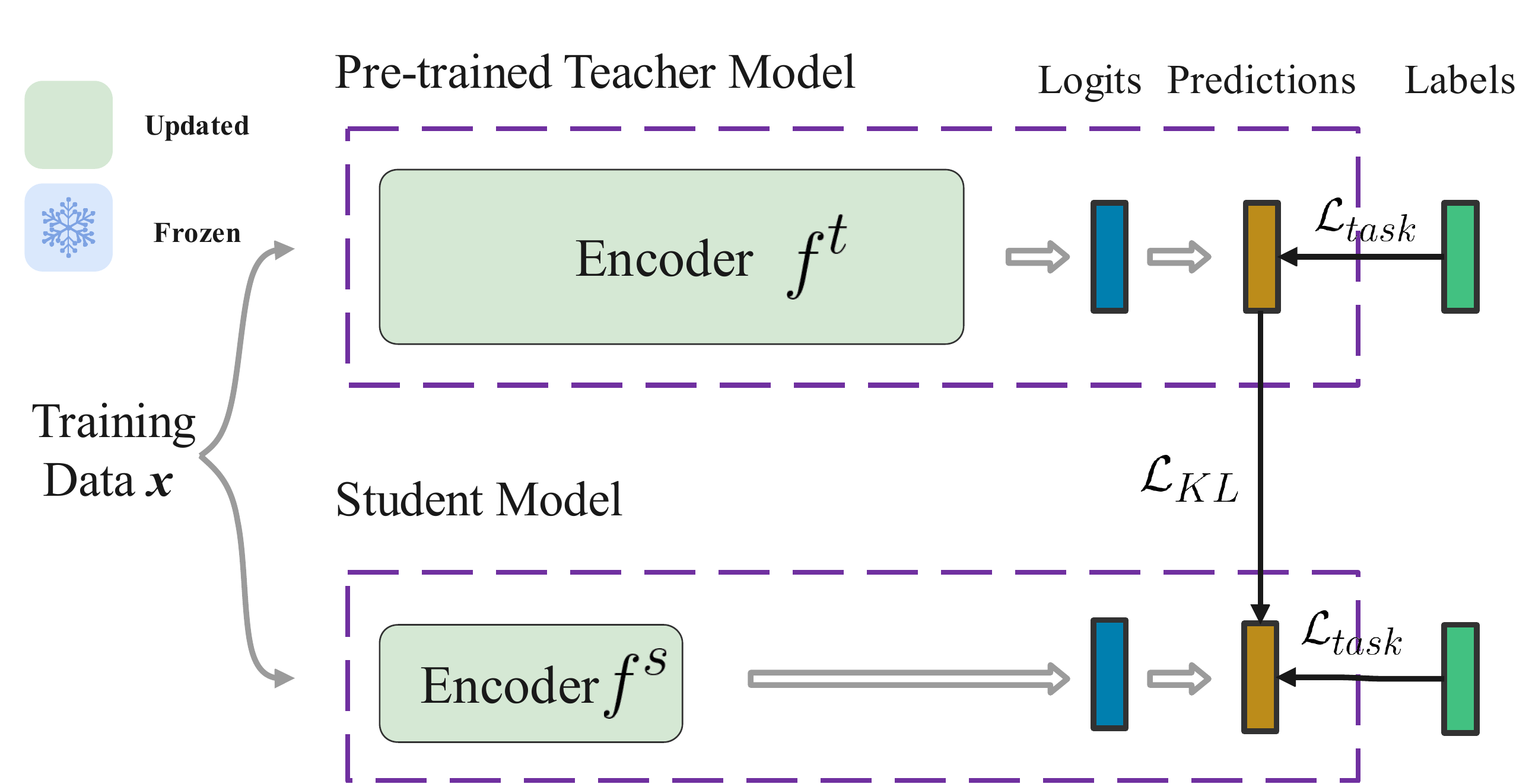}
            \label{fig:fine_tune_kd}
    \end{minipage}
    }
    \subfigure[PESF-KD]{
    \begin{minipage}[t]{0.31\textwidth}
            \centering
            \includegraphics[width=\textwidth]{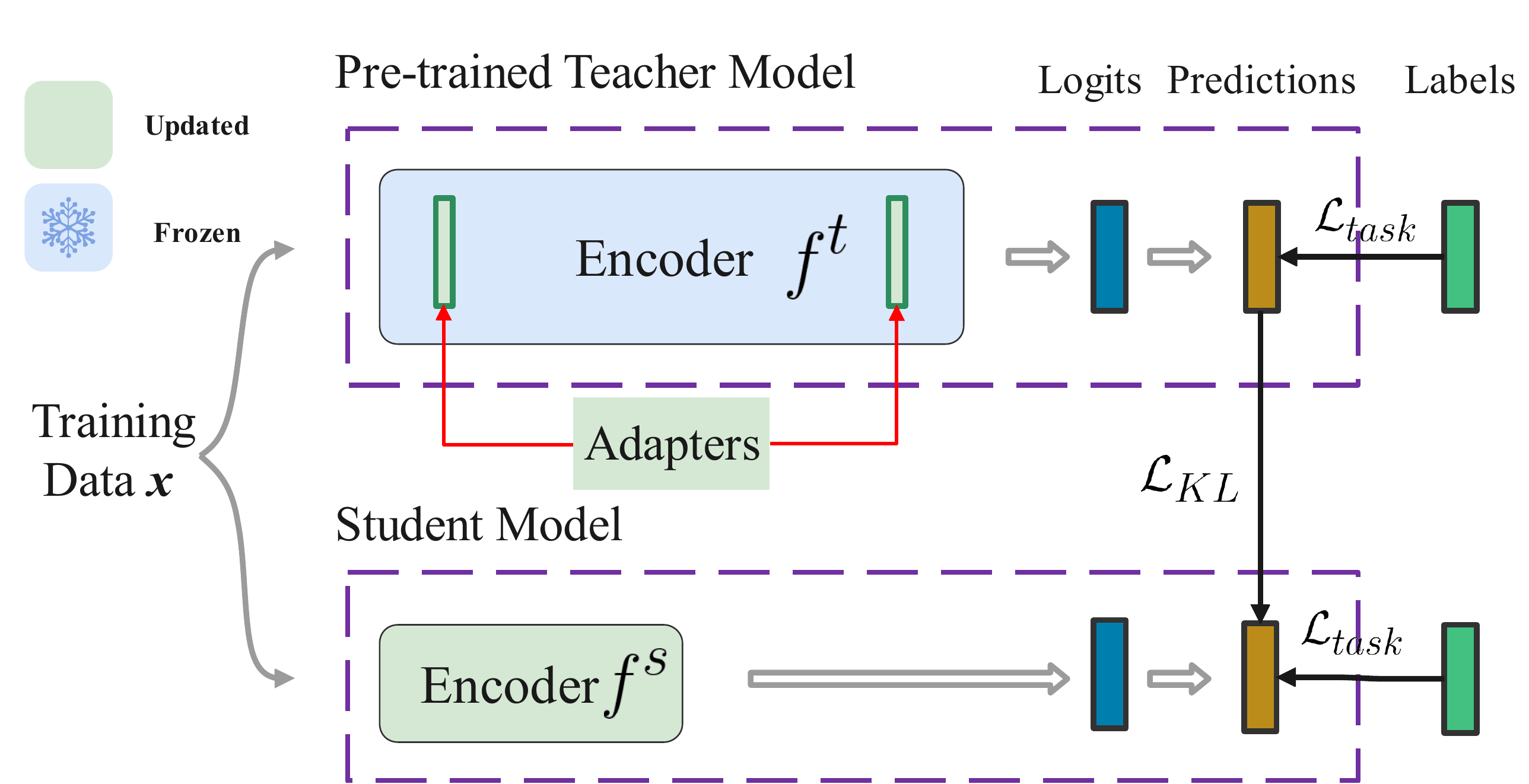}
            \label{fig:adapter_kd}
    \end{minipage}
    }
    \caption{{Comparison between KD, KD w/ fine-tuned teacher, and our PESF-KD. \textbf{\textcolor[RGB]{160,255,232}{Green}} means the parameter needs to be updated, while \textbf{\textcolor[RGB]{160,232,255}{blue}} means not.} (a) Vanilla KD, teachers and students are trained independently, resulting a gap of knowledge transfer. (b) This method is quite like DML~\cite{DML} and other online KD methods, which get better knowledge transfer and need training teachers and students together. (b) PESF-KD updates the parameters of adapter modules of the teacher with ground-truth labels and the feedback from student outputs, while the rest of the parameters of the teacher are all fixed. This method makes knowledge transfer more friendly and effective.
    }
\end{figure*}

Knowledge distillation (KD) \cite{DBLP:journals/corr/HintonVD15}, as an important method for model compression, has been widely used in various fields \cite{meta-kd,RCO,CRD,Zhang2022FinetuningGM} of deep learning. This paradigm utilizes a pre-trained teacher network to obtain a student network that is close to the teacher network but with fewer parameters. In the traditional distillation framework \cite{CRD,DBLP:conf/eccv/PassalisT18/PKT,DBLP:conf/cvpr/ParkKLC19/RKD}, the prediction output (soft label) is produced by the 
fixed teacher, as shown in Figure \ref{fig:base_kd} .
However, limited by the capacity of the student model, the teacher's knowledge cannot be well transferred to such small models, which makes traditional distillation methods suboptimal \cite{RCO,DBLP:conf/aaai/MirzadehFLLMG20}. Based on this phenomenon, we propose the first research question: \textit{\textbf{RQ1:} How to transfer teachers' knowledge to students more effectively?}

In recent works \cite{park2021learning,RCO,BERT-of-Theseus,DBLP:conf/cikm/RaoQQW0021}, the teacher network and the student network are jointly trained to make the teacher's knowledge more friendly to the students. The online distillation methods, such as DML \cite{DML} and KDCL \cite{KDCL}, usually update most or even all the parameters of the teacher by using the real labels and the feedback information (soft labels) of the students, as shown in Figure ~\ref{fig:fine_tune_kd}. However, in the setting of online distillation, a large teacher network and a student network need to be trained simultaneously for each new downstream task from scratch, which is too time-consuming and inefficient. This  motivates us to consider the following research question: \textit{\textbf{RQ2:} How to make the friendly transfer of knowledge more parameter-efficiently?}

In this paper, we propose a new distillation framework that is parameter-efficient and student-friendly, as shown in Figure~\ref{fig:adapter_kd}. 
Our paper forges a connection between two literatures that have evolved  independently: knowledge distillation and parameter-efficient learning. This allows us to leverage the powerful methods of parameter-efficient learning to ensure the friendliness of teachers' knowledge to students, and to update teachers' knowledge representation more efficiently. 
Based on extensive experiments and analyses, we show that our framework can utilize the information from ground-truth labels and student supervision to train the adapter modules, and further narrow the gap between the teacher and student models, which makes knowledge transfer easier.

In summary, our contributions are:
\begin{itemize}[leftmargin=12pt]
    \item We propose a parameter-efficient and student-friendly distillation (PESF-KD) framework, where an adapter is deliberately designed for the teacher.
    \item We provide theoretical and experimental evidences to show PESF-KD can facilitate the knowledge transfer by reducing the gap between teacher and student.
    \item We empirically validate the effectiveness and efficiency of our PESF-KD upon several vision and language models compared to existing knowledge distillation methods.  
\end{itemize}

\section{Background}
\subsection{Label Smoothing and Knowledge Distillation}
Gradient descent is mostly used to optimize the cross entropy \cite{DBLP:conf/nips/BaumW87} of the hard ground-truth labels and predicted values in deep learning. And in recent years, \citet{DBLP:conf/cvpr/SzegedyVISW16/label_smoothing} introduced \textbf{Label Smoothing} to soften and weight the traditional hard labels with the uniform distribution. 
This approach has successfully improved the effectiveness of several deep learning models and has been widely validated in natural language processing (NLP) and computer vision (CV). And to date, this approach has also been used as a training trick to improve the training of models. We provide a mathematical description of the label smoothing process. First we show the original cross-entropy:
\begin{equation}\label{eq1}
H(\boldsymbol{y}, \boldsymbol{p})=\sum_{k=1}^{K}-y_{k} \log \left(p_{k}\right),
\end{equation}
where $y_k$ is "1" for the correct class and "0" for the rest.
Then the label smoothing is achieved by increasing the smoothing parameter $\alpha$ to change $y_k$ to $y_k^{LS}$:
\begin{equation}\label{eq2}
    y_{k}^{L S}=y_{k}(1-\alpha)+\alpha / K
\end{equation}
When a network is trained with label smoothing, the differences between the logits of the correct and incorrect classes become a constant that is dependent on $\alpha$, while knowledge distillation provides dynamic soft labels to let the network learn the distribution of teachers.

\textbf{Knowledge Distillation} (KD) \cite{DBLP:journals/corr/HintonVD15} often employs a pre-trained teacher network with the goal of transferring the teacher's knowledge to a small group of students, as shown in Figure \ref{fig:base_kd}. In the classification task, one of the simplest forms is to provide the soft label information by forwarding the teacher's output. The initial teacher and student model can be defined as: teacher $\boldsymbol{p}(\theta^t)$ and student $\boldsymbol{p}(\theta^s)$, respectively, where $\theta$ is the model parameters and $\boldsymbol{p_k}(\cdot)=\frac{\exp \left(\boldsymbol{z}_{k}(\theta) / \tau\right)}{\sum_{j=1}^{K} \exp \left(\boldsymbol{z}_{j}(\theta) / \tau\right)}$ is the probability predict of the matching label and $K$ is the number of classes and $z_k$ is the logical output of the $k\mathrm{-th}$ class. So the vanilla KD loss measuring the KL-Divergence of teachers and students can be formulated as:
\begin{equation}\label{eq3}
\mathcal{L}_{K L}\left(\boldsymbol{p}(\tau|\theta^s), \boldsymbol{p}(\tau|\theta^t)\right)=\tau^{2} \sum_{j} \boldsymbol{p}_{j}(\tau|\theta^t)\cdot\\\log \frac{\boldsymbol{p}_{j}(\tau|\theta^t)}{\boldsymbol{p}_{j}(\tau|\theta^s)}
\end{equation}
where $\tau$ is the temperature used in KD, which controls how much to rely on the teacher’s soft predictions. For simplicity, we use $\mathcal{L}_{K L}$ represent $\mathcal{L}_{K L}\left(\boldsymbol{p}(\tau|\theta^s), \boldsymbol{p}(\tau|\theta^t)\right)$ in the following sections.

\subsection{Parameter-Efficient Training}
Transfer learning from pre-trained models is a general learning paradigm that has been applied to a variety of tasks \cite{DBLP:conf/cvpr/HeZRS16/resnet,bert,vilbert}. The most popular transfer learning method nowadays is to fine-tune all of the model's parameters (full fine-tuning). But full fine-tuning is relatively expensive because parameter adjustment correlates to the need to completely retrain the entire model for each downstream task. Artificially constructed modules with a modest number of parameters are used in approaches like Prompt \cite{DBLP:conf/emnlp/LesterAC21/prompt} and Adapters \cite{DBLP:conf/icml/adapter,DBLP:conf/acl/LiL20/prefix} to achieve a fit between the pre-trained model and the downstream task with parameter-efficient training. 


\section{Methodology}\label{sec:method}
\subsection{Knowledge Distillation with Fine-Tuned Teacher}
\textbf{Motivation.}
Compared with label smoothing, knowledge distillation can improve the training of the network for the following two reasons: First, teachers can understand the nuances of different classes, and such inter-class information brings more information than label smoothing and helps students generalize some unseen data. Second, the soft distribution of teachers constrains students' learning directions such that they will not be mistakenly overconfident. For a student network, the information it learns is obtained through this soft label.
Previous works~\cite{park2021learning,DBLP:conf/iccv/ZhuW21a,RCO,DBLP:conf/aaai/MirzadehFLLMG20} pointed out that the mismatch of teacher and student network capacity can cause the \textit{knowledge transfer difficulty} of such soft label~\cite{DBLP:journals/ijcv/GouYMT21}. One of the solutions to reduce transfer difficulty is smoothing the teacher's output by adjusting the temperature \cite{guo2022reducing}.
\citet{DBLP:conf/nips/MullerKH19} indicates that the label smoothness of the target provided by the teacher exerts a great influence on the student network, and the difference information between classes determines whether the student's performance can be improved.
But manual conditioning of the label smoothness by the temperature is quite difficult and may cause the loss of inter-class information when the temperature is too high.
In our preliminary experiments, we found that if the teacher network continues to fine-tune through the ground truth labels with the rest of the settings as the valina KD~\cite{DBLP:journals/corr/HintonVD15}
, teacher's labels will be smoother, and the accuracy of the distilled student network is better, where we leave this competitive setting as the strong benchmark, namely ``\textbf{KD w/ fine-tuned teacher}'', in the main experiments (Section~\ref{exp}).
This phenomenon urges us to explore the relationship between the teacher network and the student network in Section~\ref{kd_relation}.
\\
\textbf{Training Objectives.}
As shown in Figure \ref{fig:fine_tune_kd}, different from vanilla KD~\cite{DBLP:journals/corr/HintonVD15}, our distillation method requires fine-tuning the parameters of the teacher network. The teacher network needs to output soft labels to supervise the student network, it also requires ground-truth labels to training itself.
Take classification task as instance, the corresponding teacher’s loss is:
\begin{equation}\label{eq4}
\mathcal{L}_{t}=\mathcal{L}_{task}(\theta^t)=-\sum_{i\in|X|} \sum_{c\in C}
\left[\mathbf{1}\left[\mathbf{y}_{i}=c\right]
\cdot\log p\left(\mathbf{y}_{i}=c \mid\mathbf{x}_{i};
\mathbf{\theta}^{t}\right)\right]  ,
\end{equation}
where $c$ is a class label and $C$ denotes the set of class labels.

In vanilla KD~\cite{DBLP:journals/corr/HintonVD15}, students receive soft label supervision (in Equation~\ref{eq4}) from the teacher as well as label supervision as the following formulation:
\begin{equation}\label{eq5}
\mathcal{L}_{s}=\alpha\mathcal{L}_{KL}+(1-\alpha)\mathcal{L}_{task}(\theta^s),
\end{equation}
where the task loss $\mathcal{L}_{task}$ follows the same format as the teacher network.

\subsection{Parameter-Efficient and Student-Friendly Knowledge Distillation}
\textbf{Motivation.}
As with the same training objective in the previous section, we still need to update the parameters of the teacher network, as shown in Figure \ref{fig:adapter_kd}. Inspired by the method of parameter efficient fine-tuning~\cite{DBLP:conf/icml/adapter}, we introduce the adapter structure only adjusting this part of the parameters and fixing the original teacher network parameters. We take advantage of this approach to achieve parameter efficient and student-friendly fine-tuning.
\\
\textbf{Adapter Module.}
Our PESF-KD has three key properties: (i) it allows teachers and students to learn online while maintaining good performance, (ii) it can automatically update the teacher's output based on the student's output, which reduces the sharpness of teachers (see \S \ref{kd_relation}), and (iii) it adds extra parameters trainable to the pre-trained teacher network while all parameters of the original are fixed, reducing the over-consumption of the fully trained teacher network.
Many online KD methods require retraining more than one teacher networks, so it is desirable to participate in training a teacher network with only a small number of parameters.
To achieve these properties, we propose to adopt a standard adapter module for knowledge distillation.\\
\textbf{Structures.}
Most adapters can be written as proj\_down $\rightarrow$ nonlinear $\rightarrow$ proj\_up architecture. Specifically, the adapter firstly projects the input $h$ to a lower-dimensional space with dimension $r$, utilizing a down-projection weight matrix $\boldsymbol{W}_{\text {down }} \in \mathbb{R}^{d \times r}$.
Then through a nonlinear activation function and then through a up-projection function with weight matrix $\boldsymbol{W}_{\text {up }} \in \mathbb{R}^{r \times d}$ to increase the dimension to the original dimension. Usually, these modules use a residual connection, and the final form is as follows:
\begin{equation}\label{eq6}
\boldsymbol{h} \leftarrow \boldsymbol{h}+f\left(\boldsymbol{h} \boldsymbol{W}_{\text {down }}\right) \boldsymbol{W}_{\text {up }}
\end{equation}


\section{A Closer Look at Teacher-Student Relationship in Distillation}\label{kd_relation}
\subsection{Gap Between Teacher and Student}
The relationship between the teacher network and the student network in KD is rarely discussed. When independently training the model from scratch, the larger model is more likely to output sharper values and obtain better accuracy, while the smaller model is more likely to output smoother values and obtain poorer accuracy ~\cite{DBLP:conf/cvpr/Chen0ZJ21,DBLP:conf/iccv/ZhuW21a,DBLP:conf/aaai/MirzadehFLLMG20,DBLP:conf/iccv/ChoH19}. This phenomenon (i.e., capacity mismatch) significantly exacerbates the difficulty of knowledge transfer in knowledge distillation. An optional solution to measure the difference in the sharpness of the network output between teachers and students is to directly use the entropy of the network output to get the output value of the maximum confidence score.
Differently, we use a simple and intuitive sharpness metric \cite{guo2022reducing} to
get a smooth approximation to the maximum function considering the overall information of each class. 
If we use $K$ to denote $K$ classes, the \textbf{Sharpness} is defined as the logarithm of the exponential sum of logits:
\begin{equation}\label{eq7}
    S_{sharpness}=log\sum_{j}^{K} \exp{z_j(\theta)}
\end{equation}
For the difference between teacher and student networks, we use the sharp gap between the two networks as a metric:
\begin{equation}\label{eq8}
    G_{gap}=log\sum_{j}^{K} \exp{(z_j(\theta^t))} - log\sum _{j}^{K}\exp{(z_j(\theta^s))}
\end{equation}

\subsection{How to Narrow the Gap?}
\paragraph{Mathematical Analysis}
As shown in Equation \ref{eq8}, we get the expression for the gap between teachers and students. 
In this section, we will explore what factors affect this gap. We first approximate this expression using a Taylor second expansion:
\begin{equation}\label{eq9}
\begin{aligned}
G_{ g a p}=& log\sum_{j}^{K} \exp{(z_j(\theta^t))} - log\sum _{j}^{K}\exp{(z_j(\theta^s))} \\
\approx & \log \left(K+\sum_{j}^{K} z_{j}(\theta^t)+\frac{1}{2} \sum_{j}^{K} z_{j}(\theta^t)^{2}\right) -\log \left(K+\sum_{j}^{K} z_{j}(\theta^s)+\frac{1}{2} \sum_{j}^{K} z_{j}(\theta^s)^{2}\right)
\end{aligned}
\end{equation}
Following Hindon's assumption~\cite{DBLP:journals/corr/HintonVD15} and also through experimental phenomena~\cite{guo2022reducing}, it can be known that the logits of each training sample are approximately zero-meaned so that $\sum_{j}^{K} z_{j}(\theta^s)=\sum_{j}^{K} z_{j}(\theta^t)=0$. So the gap can be rewritten as:
\begin{equation}\label{gap_explain}
\begin{aligned}
G_{ g a p} &=\log \left(K+\frac{1}{2} \sum_{j}^{K}\left(z_{j}(\theta^t)\right)^{2}\right)-\log \left(K+\frac{1}{2} \sum_{j}^{K}\left(z_{j}(\theta^s)\right)^{2}\right) \\
&=\log \left(1+\frac{1}{2 K} \sum_{j}^{K} z_{j}(\theta^t)^{2}\right)-\log \left(1+\frac{1}{2 K} \sum_{j}^{K} z_{j}(\theta^s)^{2}\right)\\
&=\log \left(1+\frac{1}{2} *\sigma_{t}^{2}\right)-\log \left(1+\frac{1}{2} *\sigma_{s}^{2}\right),
\end{aligned}
\end{equation}
where the $\sigma^{2}=\frac{1}{K}\sum_{j}^{K} z_{j}(\theta)^{2}$ is the variance of logits.

Naturally, assuming that the temperature $\tau$ is fixed. Then the change of the gap comes from the change in the variance of teachers' logits and students' logits. Once the these logits become smooth then the corresponding variance becomes smaller, if the logits become sharp then the variance becomes larger. The smoothness of the final logits results in a change in the gap. In the following sections, we compared three distillation methods, namely, vanilla KD (vanilla) with different temperature and our proposed methods -- fine-tuned teacher (train with finetune teacher) and adaptive teacher (train with adapter) to check out how temperature and respective methods affect the gap. 

\begin{figure*}[h!]
\centering
\begin{minipage}[t]{0.40\linewidth}
    \centering
    \includegraphics[width=\linewidth]{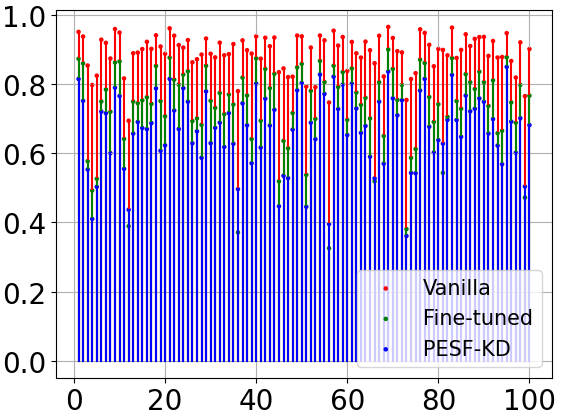}
    \caption{Normalized major logits distribution.}
    \label{major_logit}
\end{minipage}
\hspace{0.5em} 
\begin{minipage}[t]{0.40\linewidth}
\centering
    \includegraphics[width=\linewidth]{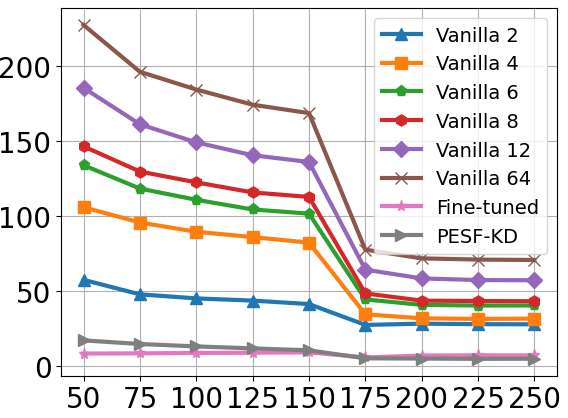}
    \caption{Sharpness gap comparison.}
    \label{sharpness_gap}
\end{minipage}\\
\hspace{0.5em}
\begin{minipage}[t]{0.88\linewidth}
\centering
    \includegraphics[width=\linewidth]{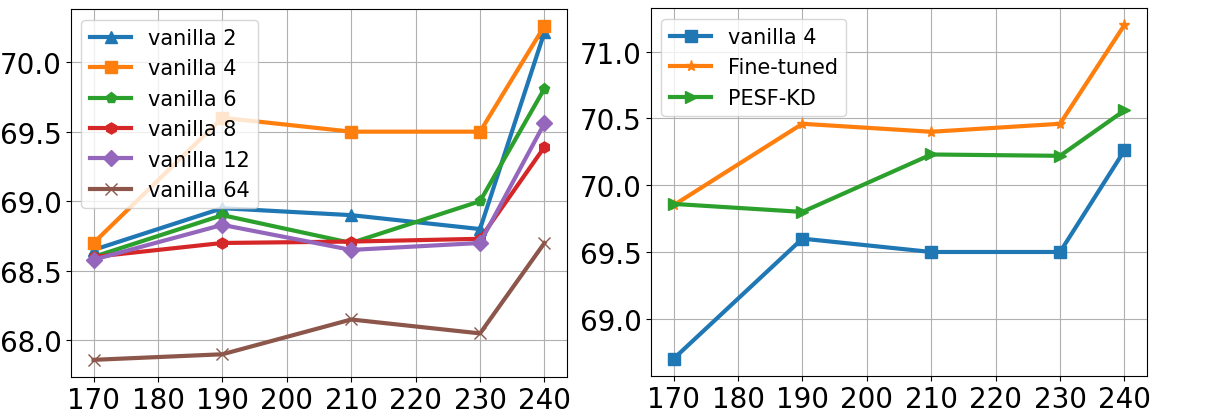}
\caption{Comparison of Top-1 accuracy between different smoothing setting and our approaches.}
\label{teacher_acc_cmp}
\end{minipage}

\end{figure*}

\begin{figure}[]
\centering
\small
\begin{minipage}[t]{0.22\linewidth}
    \centering
    \includegraphics[width=1\linewidth]{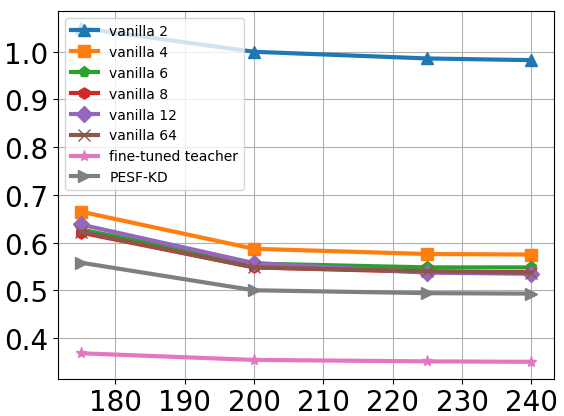}
    \caption{kl loss comparison of our proposed approach and vanilla KD.  
    }
    \label{kl_loss}
\end{minipage}
\hspace{1em}
\begin{minipage}[]{0.22\linewidth}

    \centering
    \includegraphics[width=1\linewidth]{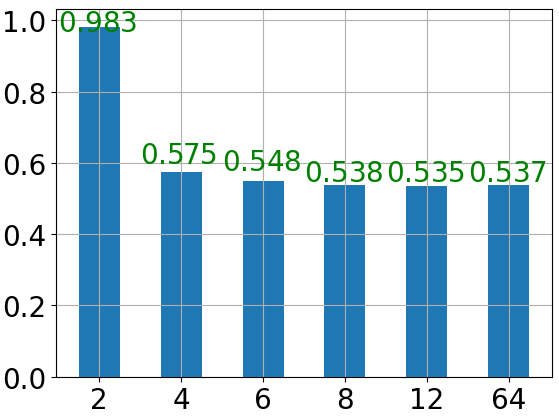}
    \caption{kl loss comparison of vanilla KD with different temperature.
    }
    \label{kl_loss_t}
\end{minipage}
\hspace{1em}
\begin{minipage}[]{0.22\linewidth}
    \centering
    \includegraphics[width=\linewidth]{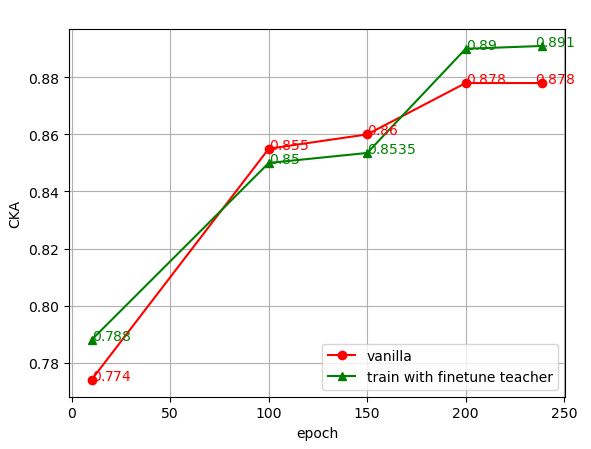}
    \caption{CKA of feature ( last layer before classifier ).  
    }
    \label{CKA_feature}
\end{minipage}
\hspace{1em} 
    \begin{minipage}[]{0.22\linewidth}
    \includegraphics[width=\linewidth]{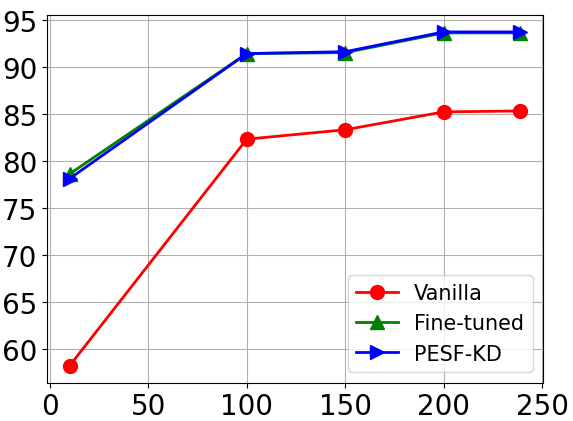}
    \caption{CKA of logit vanilla KD and ourproposed approach
    }
    \label{CKA_logit}
    \end{minipage}
\end{figure}

\paragraph{Appropriate smoothness can improve the friendliness of knowledge transfer.}
In order to more intuitively show the effect of logits output smoothness on the gap, we show the average major logit distribution in the Figure \ref{major_logit}. 
In most cases, the logit has the largest value that represents the model's category prediction, and other tiny values show that the input image is similar to those other categories.
In the real situation, the variance of students' logits will be affected by the variance of teachers' logits, and these two statistics change simultaneously. It is difficult to quantitatively analyze the gap between a fine-tuned teacher and an adaptive teacher, but we can get inspiration from Figure \ref{major_logit} and \ref{sharpness_gap}.
Obviously,  vanilla KD brings more sharp logit output, that is, greater variance. The logit output of fine-tuned teacher is in the middle position compared other two methods while PESF-KD has the smallest variance. It is worth noting that our methods greatly reduce the variance of logits compared to the vanilla KD, and both our proposed methods have relatively small variance.  
In Figure \ref{sharpness_gap}, it shows the degree of sharpness for different temperatures in vanilla KD and our methods. Consistent with the analysis of Equation \ref{gap_explain}, since the teacher's output is unchanged, that is, the variance is unchanged, by increasing the temperature, the student's output is indeed smoother, resulting in a larger gap.
Figure \ref{teacher_acc_cmp} shows the Top-1 accuracy of the corresponding methods. In general, lower gaps are associated with higher accuracy (lower temperature, lower gap value), but very close gaps may introduce anomalies (see the case of temperature 2 and 4). This phenomenon shows that smoothing the output of teachers and students within a certain range can improve the friendly transferability of knowledge (in terms of improving the accuracy and reducing the gap). In the face of each sample, it is impossible to simply manually adjust the common temperature to achieve a dynamic smooth output to improve the students' ability to accept the teacher's knowledge. With our methods, the smoothness of the output can be dynamically adjusted  according to the gradient of each sample to achieve better knowledge transfer (compared to the vanilla KD with the best temperature setting, our methods have improved a lot about accuracy, and a large reduction in the gap).
\paragraph{Student-friendly knowledge leads to better performance.}
In this part, we show that trainable part in teacher models can narrow the gap between student and teacher and make knowledge transfer process more student-friendly thus achieving better accuracy. 
We further explore the relationship between network consistency (sharpness gap in Figure \ref{sharpness_gap}, KL-Divergence in Figure \ref{kl_loss} \&  \ref{kl_loss_t} and CKA~\cite{CKA} in  Figure \ref{CKA_feature} \& \ref{CKA_logit} ) and accuracy in  Figure \ref{teacher_acc_cmp}. The three metrics mentioned above measure the final degree of consistency of the teacher-student network from different perspectives. A lower sharpness gap represents a closer knowledge representation of the teacher-student, and a lower KL represents the final convergence degree of the lower bound through distillation learning, while the a larger CKA represents larger similarity of the students and teachers.
We get the following interesting findings: 
1) From Equation \ref{gap_explain}, it is clear that the gap also decreases when the student network trained with the vanilla KD. This reduction comes from the fact that the output of the student network becomes sharper i.e., more similar to the output of the larger teacher network (both kl-divergence and gap decrease).
2) Both our proposed methods can reduce the gap, and model trained with fine-tuned teacher can bring a great reduction (from 36.3 to 16.8).
Our methods also make the output of teachers and students more consistent (the KL-Divergence, gap of the two methods are significantly lower and the CKA is higher than those of vanilla KD with different temperature).
3) The accuracy of the final student network trained by our two methods is similar, although the gap between them is quite different (32.4 vs. 16.8).
The KL-Divergence (0.42 vs. 0.23), CKA of logits and predictions (almost same) of these two proposed methods are closer, which shows that our methods can guarantee the consistency of teacher and student characteristics. 
It also illustrates that gap and KL-Divergence interpret the similarity of output distributions from different perspectives.

\begin{figure*}[t]
\centering
\subfigure[Vanilla KD]{
\begin{minipage}[]{0.3\linewidth}
    \centering
    \includegraphics[width=\linewidth]{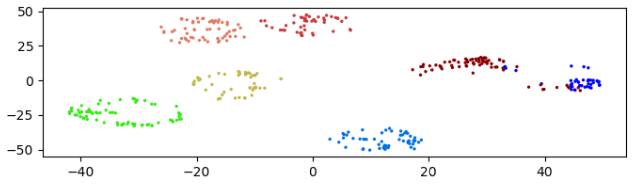}
    \label{visulization of vanilla KD}
\end{minipage}
}
\hspace{0.5em}
\subfigure[Fine-tuned teacher]{
\begin{minipage}[]{0.3\linewidth}
    \centering
    \includegraphics[width=\linewidth]{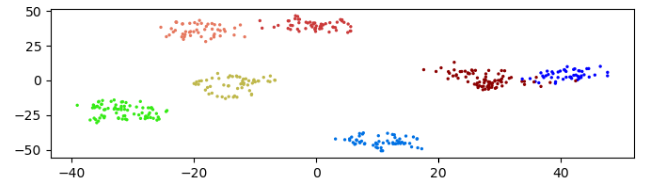}
    \label{visulization of fine-tuned teacher}
\end{minipage}
}
\hspace{0.5em}
\subfigure[PESF-KD]
{\begin{minipage}[]{0.3\linewidth}
    \centering
    \includegraphics[width=\linewidth]{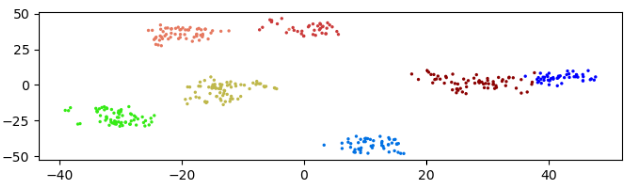}
    \label{visulization of adaptive teacher}
\end{minipage}
}
\caption{Visualize of penultimate layer  representation of Vanilla KD, w/ fine-tuned teacher and PESF-KD with 7 semantically different classes using T-SNE~\cite{van2008visualizing}. 
\textbf{Clusters are more tighter and distinguishable in our proposed approach}\label{tsne}
}
\end{figure*}

To explore the changes in class distribution of logits, we visualize the penultimate layer representation of student model ResNet20 on dataset CIFAR-100 as shown in Figure \ref{tsne}. We randomly choose 7 classes and record the penultimate layer representation in the 230-th epoch and then use T-SNE to project the data in the 2D plane. Classes are marked with different colors.
We observe that clusters in our proposed approach are tighter because the student model is encouraged to learn more information from all other class templates in the training data set by narrowing the sharpness gap between teacher and student networks.
Besides, when looking at the projections, some clusters, i.e., crimson and dark blue ones, are more discernible in our proposed models than in Vanilla KD. 

\section{Experiments}
\label{exp}

\subsection{Datasets and Baselines}
\textbf{Datasets} Two types of tasks incluing image classification (CIFAR-100~\cite{cifar100} and ImageNet~\cite{DBLP:conf/cvpr/DengDSLL009/imagenet}) and natural language understanding (GLUE~\cite{DBLP:conf/iclr/WangSMHLB19/glue} ) are adopted for a series of experiments.
Specifically, we test on two types of NLU datasets according to the amount of training resource.
\textit{\textbf{Low resource}}: MRPC~\cite{mrpc} and STS-B~\cite{sts-b} for Paraphrase Similarity Matching; SST-2~\cite{sst-2} for Sentiment Classification ; RTE~\cite{DBLP:conf/iclr/WangSMHLB19/glue} for the Natural Language Inference.
\textit{\textbf{High Resource}} :QNLI~\cite{qnli} for the Natural Language Inference and QQP
\footnote{https://www.quora.com/q/quoradata/First-Quora-Dataset-Release-Question-Pairs} 
for Paraphrase Similarity Matching.
\\
\textbf{Baselines} 
We report several knowledge distillation methods for comparison, including vanilla KD~\cite{DBLP:journals/corr/HintonVD15}, knowledge distillation via collaborative learning (KDCL) \cite{KDCL}, deep mutual learning (DML) ~\cite{DML}, contrastive representation distillation (CRD)~\cite{CRD}, relational knowledge distillation (RKD)~\cite{DBLP:conf/cvpr/ParkKLC19/RKD} and probabilistic knowledge transfer(PKT) ~\cite{DBLP:conf/eccv/PassalisT18/PKT}. 
According to ~\citet{DBLP:journals/ijcv/GouYMT21}, KD methods can be divided into two groups, online distillation and offline distillation. 
For a more fine-grained comparison, We further split them into three different kinds, online KD (DML, KDCL), offline KD (vanilla KD, PKT) and representation KD (CRD, RKD). 
Besides these methods, we report the result of our method ``w/ Fine-tuned teacher'' and ``PESF-KD'' to support our argument about less sharpness gap helps student to perform better to absorb the knowledge of the teacher.


\begin{table*}[t!]
\small
\caption{Results on CIFAR-100 test set. 
We compare the training time consumption (batch time) and the amount of parameters  needed to update (params) and the corresponding accuracy (@1) on various teacher-student combinations. The best results are \textbf{bold}. The second best results are \underline{underlined}.}
\vspace{1em}
\label{result_cifar100}
\resizebox{\textwidth}{!}
{
\begin{tabular}{@{}cccccccccccccc@{}}
\toprule
                                   &                                                               & \multicolumn{3}{c}{resnet56-20} & \multicolumn{3}{c}{resnet110-32} & \multicolumn{3}{c}{vgg13-8}  & \multicolumn{3}{c}{resnet56-vgg8} \\ \midrule
                                   & method                                                        & batch time   & params  & @1     & batch time   & params   & @1     & batch time & params & @1     & batch time   & params    & @1     \\ \midrule
\multirow{2}{*}{offline KD}        & Vanilla KD\cite{DBLP:journals/corr/HintonVD15}                                                    & 0.219        & 1.11M   & 70.95  & 0.319        & 1.89M    & 73.08  & 0.097      & 15.86M & 73.36  & 0.134        & 15.86M    & 73.98  \\
                                   & PKT\cite{DBLP:conf/eccv/PassalisT18/PKT}                                                           & 0.225        & 1.11M   & 71.27  & 0.336        & 1.89M    & 73.67  & 0.120       & 15.86M & 73.40   & 0.139        & 15.86M    & 74.10   \\ \midrule
\multirow{2}{*}{Representation KD} & CRD\cite{CRD}                                                           & 0.234        & 1.18M   & 71.44  & 0.353        & 1.96M    & 73.62  & 0.139      & 16.39M & 73.31  & 0.159        & 15.93M    & 74.06  \\
                                   & RKD~\cite{DBLP:conf/cvpr/ParkKLC19/RKD}                                                           & 0.234        & 1.11M   & 71.47  & 0.846        & 1.89M    & 73.53  & 0.207      & 15.86M & \underline{74.15}  & 0.444        & 15.86M    & 73.35  \\ \midrule
\multirow{2}{*}{online KD}&KDCL~\cite{KDCL}&0.520&4.56M& 70.11&1.035&8.84M&72.87& 0.190 & 53.71M & 73.99&0.433&19.31M& 73.16  \\
& DML\cite{DML}& 0.524        & 4.56M& 69.40  & 0.912        & 8.84M& 72.21  & 0.190       & 53.71M& \textbf{74.18}  & 0.437        & 19.31M & 73.86  \\ \midrule
\multirow{2}{*}{ours (online KD)}               & \begin{tabular}[c]{@{}c@{}}w/ Fine-tuned \\ teacher\end{tabular} & 0.481        & 4.56M &\textbf{71.65}   & 0.819        & 8.84M& \textbf{73.90}    & 0.156      & 53.71M& 73.52  & 0.443        & 19.31M &\textbf{74.40}    \\
                                   & \begin{tabular}[c]{@{}c@{}}PESF-KD\end{tabular}   & 0.228        & 1.16M   & \underline{71.63}  & 0.325        & 1.94M    & \underline{73.86}  & 0.117      & 15.91M & 73.79  & 0.136        & 15.91M    & \underline{74.29}  \\    
\bottomrule
\end{tabular}}\end{table*}

\begin{table}[t!]
\small
\centering
\begin{minipage}[t]{0.7\textwidth}
\caption{Results on CIFAR-100 test set. (continue)}
\label{acc_cifar100}
\resizebox{\textwidth}{!}
{
\begin{tabular}{@{}ccccc@{}}
\toprule
teacher                     & resnet56 & resnet110 & vgg13 & resnet56 \\
student                     & resnet20 & resnet32  & vgg8  & vgg8     \\ \midrule
teacher                     & 72.34    & 74.31     & 74.64 & 79.34    \\
student                     & 69.06    & 71.14     & 70.36 & 70.36    \\ \midrule
KD\cite{DBLP:journals/corr/HintonVD15}                          & 70.66    & 73.08     & 72.98 & 73.81    \\
FitNet\cite{journals/corr/RomeroBKCGB14}                      & 69.21    & 71.06     & 71.02 & 70.69    \\
AT\cite{DBLP:conf/iclr/ZagoruykoK17}                          & 70.55    & 72.31     & 71.43 & 71.84    \\
SP\cite{DBLP:conf/iccv/TungM19}                          & 69.67    & 72.69     & 72.68 & 73.34    \\
CC\cite{DBLP:conf/iccv/PengJLZWLZ019}                          & 69.63    & 71.48     & 70.71 & 70.25    \\
VID\cite{DBLP:conf/cvpr/AhnHDLD19}                         & 70.38    & 72.61     & 71.23 & 70.30     \\
AB\cite{DBLP:conf/aaai/HeoLY019a}                          & 69.47    & 70.98     & 70.94 & 70.65    \\
FT\cite{DBLP:conf/nips/KimPK18}                          & 69.84    & 72.37     & 70.58 & 70.29    \\
FSP\cite{DBLP:conf/cvpr/YimJBK17}                         & 69.95    & 71.89     & 70.23 & 73.90     \\
w/ Fine-tuned teacher & \textbf{71.66}    & \textbf{73.89}     & \underline{73.51} & \textbf{74.40}     \\
PESF-KD          & \underline{71.64}    & \underline{73.85}     & \textbf{73.79} & \underline{74.29}    \\ \bottomrule
\end{tabular}
}
\end{minipage}
\end{table}
\begin{table}[t!]
\centering
\begin{minipage}[t]{0.9\textwidth}
\caption{Results on ImageNet test set. ResNet18 is the student model and ResNet50 is the teacher model. We report the averaged results over 3 random seeds. The best results are \textbf{bold}. The second best results are \underline{underlined}.}
\label{result_imagenet}
\centering
\resizebox{1\textwidth}{!}
{
\begin{tabular}{ccccccc}
\toprule
                                   & method            & batch time & params & @1      & KL loss & GAP   \\ \midrule
\multirow{2}{*}{offline KD}        & Vanilla KD\cite{DBLP:journals/corr/HintonVD15}        & 0.46      & 47M& 69.20    &    10.3     &      19.3 \\
                                   & PKT\cite{PKD}& 0.47      &47M& 69.69& 10.3   & 19.2  \\
\multirow{2}{*}{online KD}         & KDCL\cite{KDCL}              & 1.56       & 149M& \textbf{70.17}   & \textbf{6.7}    & 31.8  \\
                                   & DML\cite{DML}               & 1.45      & 149M    & 68.45   & \textbf{6.7}    & \textbf{14.1}  \\
\multirow{2}{*}{Representation KD} & CRD\cite{CRD}               & 0.57       &63M&69.33&10.5  &20.0 \\
                                   & RKD\cite{DBLP:conf/cvpr/ParkKLC19/RKD}&1.53&47M&69.52&9.6&20.5 \\ 
\multirow{2}{*}{ours (online KD)}&w/ Fine-tuned teacher&1.44&149M&\underline{70.00}& \textbf{6.7}&\underline{14.9}   \\
                                   & PESF-KD           & 0.47      & 54M& 69.96   & 8.9    & 18.6
                                \\
\bottomrule
\end{tabular}
}
\end{minipage}
\end{table}


\subsection{Experimental Setup}


For CV tasks, we follow previous works~\cite{CRD,meta-kd} using various combination of student \& teacher networks. Each pair of student \& teacher networks are from different capacity and architecture, including ResNet~\cite{DBLP:conf/cvpr/HeZRS16/resnet} and VGG~\cite{vggnet}. We run two types of distillations, isomorphic distillation and isomeric distillation.
For isomorphic distillation, we run three different combinations (ResNet56-ResNet20, ResNet 110-ResNet32 and VGG13-VGG8). For isomerism distillation, the results of ResNet-56 to VGG-8 are reported. For NLU tasks, we first fine-tune the pre-trained teacher (12-layers version of BERT-Base) and then train student model (6-layers version of BERT-Base) on each downstream task. 
We report \textit{Top 1 Accuracy} (@1) for image classification experiments as a network performance metric.
For NLU tasks, we report the same format as on the GLUE leaderboard.
 For MRPC and QQP we report \textit{F1}. For STS-B, we report \textit{Pearson and Spearman correlation}. For other tasks we report \textit{accuracy}. Besides, to measure network consistency between teachers and students , we report average predicted \textit{KL divergence} (KL), \textit{CKA consistency}~\cite{CKA} (CKA) and \textit{the sharpness gap}~\cite{guo2022reducing} (GAP) as similiarity metrics.

\subsection{Results}
\begin{table*}[t!]
\small
\centering
\caption{Results on the development set of GLUE with our low \& high resource division. 
We consider tasks with fewer than 100K as low-resource tasks and others are high-resource tasks. The best results are \textbf{bold}. The second best results are \underline{underlined}. We report the averaged results over 3 random seeds.
}
\label{result_GLUE}
\resizebox{1\textwidth}{!}
{
\begin{tabular}{@{}ccccccccc@{}}
\toprule
                   &          &         & \multicolumn{4}{c}{low resource}                 & \multicolumn{2}{c}{high resource} \\ \midrule
method             & \#Params & Time & \begin{tabular}[c]{@{}c@{}}MRPC\\ $(3.7K)$\end{tabular} & \begin{tabular}[c]{@{}c@{}}STS-B\\ $(5.7K)$\end{tabular} & \begin{tabular}[c]{@{}c@{}}RTE\\ $(2.5K)$\end{tabular} & \begin{tabular}[c]{@{}c@{}}SST-2\\ $(67K)$\end{tabular} & \begin{tabular}[c]{@{}c@{}}QNLI\\ $(105K)$\end{tabular}     &  \begin{tabular}[c]{@{}c@{}}QQP\\ $(364K)$\end{tabular}               \\ \midrule
BERT-base$_{teacher}$            & -        & -       & 91.6 & 90.2/89.8 & 71.4      & 93    & 91.2        & 88.5         \\
BERT-base$_{student}$            & -        & -       & 89.2 & 88.1/87.9 & 67.9      & 91.1  & 88.6        & 86.9         \\ \midrule
Vanilla KD\cite{DBLP:journals/corr/HintonVD15}            & 66M      & 1.00x   & 89.6 & \underline{88.6}/88.2   & 67.7      & 91.2  & 89.0           & 87.3         \\
RCO\cite{RCO}                & >176M      & >2.66x   & \textbf{90.5} & \textbf{88.7}/\underline{88.3} & 67.6     & 91.4  & \textbf{89.7}        & 87.4         \\
TAKD\cite{DBLP:conf/aaai/MirzadehFLLMG20}               & >132M      & >2.00x   & 89.6 & 88.2/88.0 & 
\underline{68.5}      & 91.4  & 89.6          & 87.5         \\
DML\cite{DML}                & 176M      & 2.66x   & 89.6 & 88.4/88.1 & 68.4      & \underline{91.5}  & \underline{89.6}            & 87.4         \\
SFTN\cite{park2021learning}               & >176M      & >2.66x   & \underline{89.8} & 88.4/\textbf{88.5}  & \textbf{69.4}      & \underline{91.5}  & 89.5       & 87.5         \\
PKD\cite{PKD}                & 66M      & >1.05x   & 89.4 & \underline{88.6}/88.1 & 67.6      & 91.3  & 89.5          & \textbf{87.8}        \\
w/ Fine-tuned teacher & 176M     & 2.66x   & \underline{89.8} & \underline{88.6}/\underline{88.3} & 68.2      & \textbf{91.7}  & 89.2       & 87.2         \\
PESF-KD   & 66M      & 1.05x   & 89.7 & \underline{88.6}/\underline{88.3} & 68.2      & \underline{91.5}  & 89.1       & \underline{87.7}        \\ \bottomrule
\end{tabular}
}
\end{table*}
\subsubsection{Results on CIFAR-100}
%
This section compares different KD methods, which can be grouped as offline KD, representation KD, and online KD, and their training running time (batch time), need for updated parameters (params), and the corresponding accuracy., as shown in Table \ref{result_cifar100}. 
Encouragingly, our experiments fit our two previous hypotheses.
On the one hand, through the joint training of the teacher network and the student network, the problem of difficult knowledge transfer when the teacher-student capacity does not match can be alleviated. Specifically, the distillation results of the online distillation methods are significantly higher than those of the offline distillation methods, especially in the case of excessive difference in network capacity between teachers and students (vgg13-8) get a huge improvement.
And our method ``w/ Fine-tuned teacher'' achieved the best results among almost all distillation methods, such as 71.65 for resnet56-20 and 73.90 for resnet110-32
and 74.40 for resnet56-vgg8.

On the other hand, through the adapter module, the amount of parameters that teachers need to update can be reduced on the premise of alleviating the problem of capacity mismatch, thereby reducing the cost of training.
As shown in Table \ref{result_cifar100}, the method of online distillation will greatly increase the training cost due to the need to update the parameters of the teacher network and the student network synchronously (see the batch time change).
While PESF-KD significantly reduces the number of parameters that need to be updated for training, i.e., vgg13-8 reduces the number of parameters from the $\sim$ 54M to $\sim$ 16M.
Obviously, it also significantly reduces the time required for training.
Whereas in most combinations of teacher-student architectures on distillation, even if using adaptive teacher has a slight reduction in the best distillation results, such reduction is negligible. To further illustrate the superiority of our methods, we further compare the current typical distillation methods like previous work~\cite{CRD}, as shown in Table \ref{acc_cifar100}. Results of baselines in Table \ref{acc_cifar100} are reported  by \cite{meta-kd}.
The results show that our two proposed methods achieve the best results among these methods under a broad combination of teacher-student architectures in distillation.

\subsubsection{Results on ImageNet}
We further conduct experiments for ResNet (ResNet50, 102.2M to ResNet18, 46.8M) distillation on a lager dataset ImageNet. As shown in Table 3, our methods achieves comparable Top-1 accuracy results compared to other baselines, including SOTA representation KD method CRD (70.00 for fine-tuned teacher; 69.96 for PESF-KD). Notably, compared with other online KD method KDCL (70.17), student trained with fine-tuned teacher show similar top-1 accuracy (fall behind 0.17 point) and the same KL loss. That’s partly because both the methods have similar working mechanisms which allow part of the teacher model to be trainable during the student training process. But KDCL fails to fill the sharpness gap between teacher and student (31.8), while our methods are more capable of narrowing the gap (14.9 and 19.2). Besides, the student trained with the adaptive teacher reach similar top-1 accuracy (69.96) compared with fine-tuned teacher (70.00) and only require less than one-third the time (0.47 batch time for adaptive teacher and 1.44 batch time for fine-tuned teacher). That fits our assumption of PESF-KD to improve performance and reduce training time.

\subsubsection{Results on GLUE}

In Table \ref{result_GLUE}, we further explore the effect of the PESF-KD in NLP dataset, and other baselines are reported by \cite{meta-kd}. Time in the Table \ref{result_GLUE} refer to training resources cost, which is the lowest consumption with our PESF-KD compared with other baselines except for vanilla KD.  
In \cite{zhang2022fine}, they find that the overly strict regularization of KD is unnecessary in NLP datasets, which will actually reduce the final performance using KD in most cases of GLUE. However, PESF-KD can even bring further improvements to most datasets with KD, probably due to better student learning from teachers.
Compared to vanilla  KD~\cite{DBLP:journals/corr/HintonVD15}, PESF-KD achieve consistent improvement in low-resource situations, showing that the same effect of low-resource adaptability~\cite{DBLP:conf/icml/adapter}. 
Similar to the results of CV datasets, our method also obtains similar results and requires minimal training cost compared with baselines on NLP datasets, 
and even surpasses other baselines on SST-2, 
which verifies the generalizability of our methods. 

\begin{figure*}[t]
  \begin{minipage}[t]{0.42\linewidth}
      \centering
      \includegraphics[width=\linewidth]{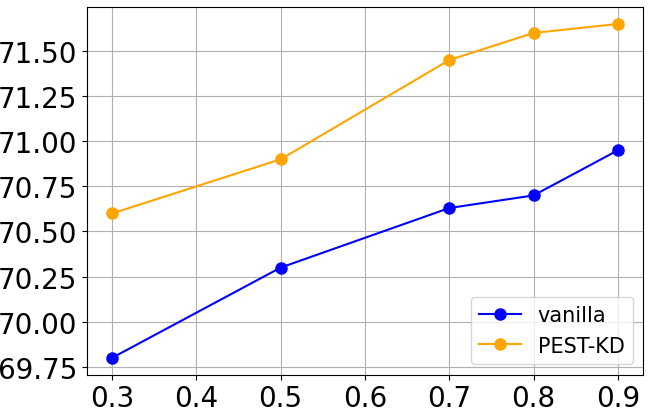}
      \caption{Influence of Loss weight on results of different KD methods.  \textbf{$\alpha$ = 0.9  is an appropriate choice.}
      \label{alpha_ablation}}
\end{minipage}
  \hspace{5em}
  \begin{minipage}[t]{0.42\linewidth}
      \centering
      \includegraphics[width=\linewidth]{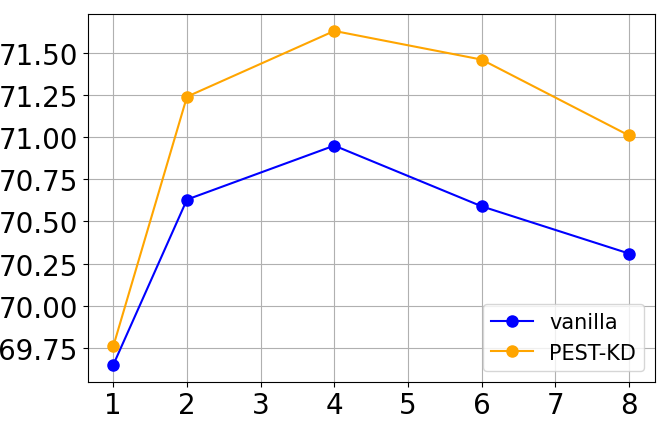}
      \caption{Influence of temperature on results of different KD methods.  \textbf{T=4 is an appropriate choice.}}
      \label{temperature_ablation}
  \end{minipage}
\end{figure*}

\begin{table}[t!]
 \small
 \centering
 \hspace{-1.4em}
\begin{minipage}[t]{0.58\textwidth}
\caption{Ablation on different structures of adapter in VGG. Experiments are performed from CIFAR-100. \textbf{The simple 3-layer adapter shows better performance compared with other parameter efficient method in top 1 accuracy, CKA and time consuming.}}
\label{vgg_ablation}
\resizebox{\textwidth}{!}
{
 \begin{tabular}{@{}ccccc@{}}
 \toprule
 method          & batch time & parameter & top-1 & CKA    \\ \midrule
 adapter \cite{DBLP:conf/icml/adapter}       & 0.137      & 15.95M    & 73.43 & 0.8758 \\
 LoRA \cite{lora}           & 0.166      & 15.95M    & 73.49 & 0.8579 \\
 Scaled PA \cite{toward_unified_view_adapter}    & 0.165      & 15.95M    & 73.13 & 0.8594 \\ \bottomrule
 \end{tabular}
}
\end{minipage}
\hspace{0.3em}
\begin{minipage}[t]{0.4\textwidth}
\caption{Comparison of different methods with PESF-KD. \textbf{It shows that PESF-KD has the potential to enhance distillation performance on other KD methods.}}
\label{ablation of different method}
\centering
\resizebox{1\textwidth}{!}
{
\begin{tabular}{@{}ccc@{}}
\toprule
teacher/student & \multicolumn{2}{c}{resnet56/resnet20} \\ \midrule
method & stadard           & PESF-KD           \\ \midrule
KD~\cite{DBLP:journals/corr/HintonVD15}              & 70.95             & 71.63             \\
PKT~\cite{DBLP:conf/eccv/PassalisT18/PKT}             & 71.27             & 71.74             \\
CRD~\cite{CRD}             & 71.44             & 71.76             \\
RKD~\cite{DBLP:conf/cvpr/ParkKLC19/RKD}             & 71.47             & 71.52             \\ \bottomrule
\end{tabular}}
\end{minipage}
\end{table}

\subsection{Analysis}

\paragraph{Adapter Architecture}
The adapter module is our recipe for success in the above performance comparisons. To explore the influence of adapter structure on classification results, we compare three classic adapter structures and report their top-1 accuracy and CKA consistency.
We use the teacher-student combination of vgg13-8 and conduct comparative experiments on CIFAR-100.
As can be seen from Table \ref{vgg_ablation}, 
the simple and efficient adapter achieve second  performance in top-1 accuracy, best CKA scores  and time consuming.
Therefore we use simple adapter module in all experiments.

\paragraph{Loss Weight} The hyper-parameter $\alpha$ decides the percent of cross entropy loss between the ground truth labels and student models prediction, and the KL loss between student and teacher models. 
Bigger $\alpha$ means a higher percent of kl loss. Figure \ref{alpha_ablation} shows that  bigger $\alpha$ benefits student models in PESF-KD. That's partly support our hypothesis of PESF-KD to promote a easy-learning teacher.

\paragraph{Temperature} Figure \ref{temperature_ablation} shows the performance of student models in differ temperature. In the experiments vanilla KD and our proposed approach has similar trend in different temperatures. And PESF-KD shows better performance in all experiments.

\paragraph{Results combined with other methods} We also test PESF-KD in other methods to verify the potential improvement on other knowledge distillation methods. Table \ref{ablation of different method} shows the results of four knowledge distillation approaches (vanilla KD~\cite{DBLP:journals/corr/HintonVD15}, PKT~\cite{DBLP:conf/eccv/PassalisT18/PKT}, CRD~\cite{CRD} and ~RKD\cite{DBLP:conf/cvpr/ParkKLC19/RKD}) combined with PESF-KD on CIFAR-100. 
Compared with standard knowledge distillation approaches, results with  PESF-KD get better performance on top-1 accuracy and indicate that PESF-KD has the potential to become a plug-in method on top of different KD methods.
\section{Conclusion}
In this paper, we present PESF-KD, a novel knowledge distillation framework by applying adapters to optimize the teacher network for better knowledge transfer to the student network. Through detailed analysis, we point out that the decline in sharpness and a better ability to distinguish within classes lead to better knowledge transfer, which leads to better results.
Extensive experiments demonstrate the robustness and effectiveness of our method.
Future works of embedding PESF-KD into existing distillation framework is expected to generalize our methods.








{\small
\bibliographystyle{acl_natbib}
\bibliography{neurips_2022}
}

\appendix

\section{Appendix}

\subsection{Training details}

\begin{table}[h]
\centering
\caption{Statistics of datasets}
\label{Statistics of dataset}
\begin{tabular}{@{}cccccc@{}}
\toprule
\multicolumn{2}{c}{dataset}      & Training set & Testing set & Development set & Classes \\ \midrule
\multicolumn{2}{c}{CIFAR-100\cite{cifar100}}    & 50000                   & 10000                  & -                  & 100               \\
\multicolumn{2}{c}{ImageNet\cite{DBLP:conf/cvpr/DengDSLL009/imagenet}}     & $\approx$1200k                  & $\approx$100k                  & $\approx$50k               & 1000              \\
\multirow{10}{*}{GLUE\cite{DBLP:conf/iclr/WangSMHLB19/glue}} & CoLA    & 8551                    & 1063                   & 1043               & 2                 \\
                       & SST-2   & 67350                   & 1821                   & 873                & 2                 \\
                       & MRPC    & 3668                    & 1725                   & 408                & 2                 \\
                       & STS-B   & 5749                    & 1377                   & 1379               & 5                 \\
                       & QQP     & 363870                  & 390965                 & 40431              & 2                 \\
                       & MNLI-m  & 390702                  & 9796                   & 9815               & 3                 \\
                       & MNLI-mm & 390702                  & 9847                   & 9832               & 3                 \\
                       & QNLI    & 104743                  & 5461                   & 5463               & 2                 \\
                       & RTE     & 2491                    & 3000                   & 277                & 2                 \\
                       & WNLI    & 635                     & 146                    & 71                 & 2                 \\ \bottomrule
\end{tabular}
\end{table}

\paragraph{Datasets} 
Table \ref{Statistics of dataset} shows the statistics used in our experiments. The training set, testing set, and development set refer to the scale of data in each part of the dataset. Table \ref{Statistics of dataset} shows that ImageNet is much bigger than CIFAR-100 in scale. And that's the reason we choose the two datasets, to verify our proposed approach in datasets of different scales. GLUE benchmark can be naturally split into low resource tasks and high resource ones for the similar propose. We divide low-resource tasks and high-resource ones by 100k on a data scale.

We are also concerned about people's consent and privacy in datasets. From the information search from the website, faces in ImageNet have been blurred in order to protect privacy after being blamed to collect information from people without consent. But we do not find a similar announcement on the website of CIFAR-100 and GLUE. Maybe it's because they have pre-processed the data or the data itself contains little privacy information. It's better for the websites to make the announcement of not containing personally identifiable information or offensive content.

\paragraph{Infrastructure} 
We implement our models with Pytorch, and our experiments are as follows:

\begin{enumerate}
\item CPU: 256  AMD EPYC 7742 64-Core Processor
\item RAM: 386840MB
\item GPU: 8x GeForce RTX 3090
\item Operating System: Ubuntu 18.04 LTS
\item Tools: Python3.7, tensorflow2,2,0, sklearn 0.23.2
\end{enumerate}

\paragraph{Hyper-parameter search}
In CV experiment, we follow previous works~\cite{CRD}, the settings on CIFAR-100 and ImageNet dataset are the same as these works. 
In CIFAR-100 we train the student model by SGD optimizer with a momentum of 0.9, a batch size of 64 and weight decay of $5\times10^{-4}$. The learning rate starts from 0.05 and decays by 10 every 30 epochs after 150 epochs. 
And on ImageNet we train the student model by SGD optimizer with a momentum of 0.9, a batch size of 256 and weight decay of $1\times10^{-4}$. The learning rate starts from 0.1 and decays by 10 every 30 epochs after 30 epochs. In the experiment of students trained with the fine-tuned teachers, we train the teacher model along with the student model during the training process with a learning rate of $1\times10^{-3}$. 
And in the experiment of a student trained with a teacher with the adapter module, also called adaptive teacher, the learning rate of the trainable part in the teacher model is set to $1\times10^{-4}$. Notably, classification loss from the teacher model is appended to the loss of students by multiplying a hyper-parameter of 0.5.
 
For NLP, we inherit parameters like maximum sequence length, temperature and batch size  according to setting from previous works~\cite{meta-kd}. We also perform grid search over the sets of the student learning rate $\lambda$ from \{1e-5, 2e-5, 3e-5, 4e-5, 5e-5\}, teacher learning rate $\mu$ from \{2e-6, 1e-5, 2e-5\}, batch size from \{32, 64\}, the weight of KD loss from \{0.3, 0.5, 0.8, 0.9\} for a better performance. We evaluate our methods on dev set of GLUE benchmark. 

\end{document}